\def\year{2018}\relax
\documentclass[letterpaper]{article} 
\usepackage{aaai18}  
\usepackage{times}  
\usepackage{helvet}  
\usepackage{courier}  
\usepackage{url}  
\usepackage{graphicx}  
\usepackage{algorithm}
\usepackage{algpseudocode}
\usepackage{bm}
\usepackage{amssymb}
\usepackage{multirow}
\usepackage{amsmath}

\def\argmax{\mathop{\rm argmax}}

\begin{document}
%
\title{Zero-Resource Neural Machine Translation with \\ Multi-Agent Communication Game}
\author{{Yun Chen$^\dagger$, Yang Liu$^\ddagger$,  Victor O.K. Li$^\dagger$}\\
	$^\dagger$Department of Electrical and Electronic Engineering, The University of Hong Kong\\ $^\ddagger$State Key Laboratory of Intelligent Technology and Systems \\
	Tsinghua National Laboratory for Information Science and Technology \\
	Department of Computer Science and Technology, Tsinghua University, Beijing, China \\
	\tt yun.chencreek@gmail.com; liuyang2011@tsinghua.edu.cn; vli@eee.hku.hk}
\maketitle
\begin{abstract}
While end-to-end neural machine translation (NMT) has achieved notable success in the past years in translating a handful of resource-rich language pairs, it still suffers from the data scarcity problem for low-resource language pairs and domains. To tackle this problem, we propose an interactive multimodal framework for zero-resource neural machine translation. Instead of being passively exposed to large amounts of parallel corpora, our learners (implemented as encoder-decoder architecture) engage in cooperative image description games, and thus develop their own image captioning or neural machine translation model from the need to communicate in order to succeed at the game. Experimental results on the IAPR-TC12 and Multi30K datasets show that the proposed learning mechanism significantly improves over the state-of-the-art methods.
\end{abstract}

\section{Introduction}
Neural machine translation (NMT) \cite{Kalchbrenner2013RecurrentCT,Sutskever2014SequenceTS,Bahdanau2014NeuralMT}, which directly models the translation process in an end-to-end way, has achieved state-of-the-art translation performance on resource-rich language pairs such as English-French and German-English \cite{Johnson2016GooglesMN,Gehring2017ConvolutionalST,Vaswani2017AttentionIA}. The success is mainly attributed to the quality and scale of available parallel corpora to train NMT systems. However, preparing such parallel corpora has remained a big problem in some specific domains or between resource-scarce language pairs. Zoph et al. \shortcite{Zoph2016TransferLF} indicate that NMT trends to obtain much worse translation quality than statistical machine translation (SMT) under small-data conditions. 

As a result, developing methods to achieve neural machine translation without direct source-target parallel corpora has attracted increasing attention
in the community recently. These methods utilize a third language \cite{Firat2016ZeroResourceTW,Johnson2016GooglesMN,Chen2017ATF,Zheng2017MaximumEL,Cheng2017JointTF} or modality \cite{Nakayama2017ZeroresourceMT} as a pivot to enable zero-resource source-to-target translation. Although promising results have been obtained, pivoting with a third language still demands large scale parallel source-pivot and pivot-target corpora. On the other hand, large amounts of monolingual text documents with rich multimodal content are available on the web, e.g., text with photos or videos posted to social networking sites and blogs. How to utilize the monolingual multimodal content to build zero-resource NMT systems remains an open question.

Multimodal content, especially image, has been widely explored in the context of NMT recently. Most of the work focus on using image in addition to text query to reinforce the translation performance \cite{Caglayan2016DoesMH,Hitschler2016MultimodalPF,Calixto2017DoublyAttentiveDF}. This task is called multimodal neural machine translation and has become a subtask in WMT16 \footnote{http://www.statmt.org/wmt16/} and WMT17 \footnote{http://www.statmt.org/wmt17/}. In contrast, there exists limited work on bridging languages using multimodal content only. Gella et al. \shortcite{Gella2017ImagePF} propose to learn multimodal multilingual representations of fixed length for matching images and sentences in different languages in the same space with image as a pivot. Nakayama and Nishida \shortcite{Nakayama2017ZeroresourceMT} suggest putting a decoder on top of the fixed-length modality-agnostic representation to generate a translation in the target language. Although the approach enables zero-resource translation, the use of a fixed-length vector is a bottleneck in improving translation performance \cite{Bahdanau2014NeuralMT}.

In this work, we introduce a multi-agent communication game within a multimodal environment \cite{Lazaridou2016TowardsMC,Havrylov2017EmergenceOL} to achieve direct modeling of zero-resource source-to-target NMT. We have two agents in the game: a captioner which describes an image in the source language and a translator which translates a source-language sentence to a target-language sentence. Apparently, the translator is our training target. The two agents collaborate with each other to accomplish the task of describing an image in the target language with message in the source language exchanged between the agents. Both agents get reward from the communication game and collectively learn to maximize the expected reward. Experiments on German-to-English and English-to-German translation tasks over the IAPR-TC12 and Multi30K datasets demonstrate that the proposed approach yields substantial gains over the baseline methods.

\section{Background}
Given a source-language sentence $\mathbf{x}$ and a target-language sentence $\mathbf{y}$, a NMT model aims to build a single neural network $P(\mathbf{y}|\mathbf{x};\bm{\theta}_{x \rightarrow y})$ that translates $\mathbf{x}$ into $\mathbf{y}$, where $\bm{\theta}_{x \rightarrow y}$ is a set of model parameters. For resource-rich language pairs, there exists a source-target parallel corpus $D_{x,y}$ to train the NMT model. The model parameters can be learned with standard maximum likelihood estimation on the parallel corpus:
\begin{eqnarray}
\hat{\bm{\theta}}_{x \rightarrow y} = \argmax_{\bm{\theta}_{x \rightarrow y}} \Bigg\{  \sum_{\langle \mathbf{x}, \mathbf{y} \rangle \in D_{x,y} } \log P(\mathbf{y}|\mathbf{x}; \bm{\theta}_{x \rightarrow y}) \Bigg\}. 
\end{eqnarray}

Unfortunately, parallel corpora are usually not readily available for low-resource language pairs or domains. On the other hand, there exists monolingual multimodal content (images with text descriptions) in the source and target language. It is possible to bridge the source and target languages with the multimodal information \cite{Nakayama2017ZeroresourceMT} for an image is a universal representation across all languages.

One way to ground a natural language to a visual image is through image captioning, which annotates a description for an input image with natural language through a CNN-RNN architecture \cite{Xu2015ShowAA,karpathy2015deep}. Below, we call a pair of a text description and its counterpart image a ``document'' and use $\mathbf{z}$ to denote an image. Given documents in the target language $D_{z,y}=\{\langle \mathbf{z}^n,\mathbf{y}^n \rangle \}_{n=1}^{N}$, an image caption model $P(\mathbf{y}|\mathbf{z}; \bm{\theta}_{z \rightarrow y})$ can be built, which ``translates'' an image to a sentence in the target language. The model parameters $\bm{\theta}_{z \rightarrow y}$ can be learned by maximizing the log-likelihood of the monolingual multimodal documents:
\begin{eqnarray}
\hat{\bm{\theta}}_{z \rightarrow y} = \argmax_{\bm{\theta}_{z \rightarrow y}} \Bigg\{  \sum_{\langle \mathbf{z}, \mathbf{y} \rangle \in D_{z,y} } \log P(\mathbf{y}|\mathbf{z}; \bm{\theta}_{z \rightarrow y}) \Bigg\}. 
\end{eqnarray}

Inspired by the idea of pivot-based translation \cite{Cheng2017JointTF,Zheng2017MaximumEL}, another way to achieve image-to-target translation is using a second language (the source language) as a pivot. As a result, image-to-target translation can be divided into two steps: the image is first translated to a source sentence using the image-to-source captioning model, which is then translated to a target sentence using the source-to-target translation model. We use two agents to represent the image-to-source captioning model and the source-to-target translation model. The image-to-target translation procedure can be simulated by a two-agent communication game, where agents cooperate with each other to play the game and collectively learn their model parameters based on the feedback. Below we formally define the game, which is a general learning framework for training zero-resource machine translation model with monolingual multimodal documents only.

\section{Two-agent Communication Game}
\begin{figure}[!t]
	\centering\includegraphics[width=3in]{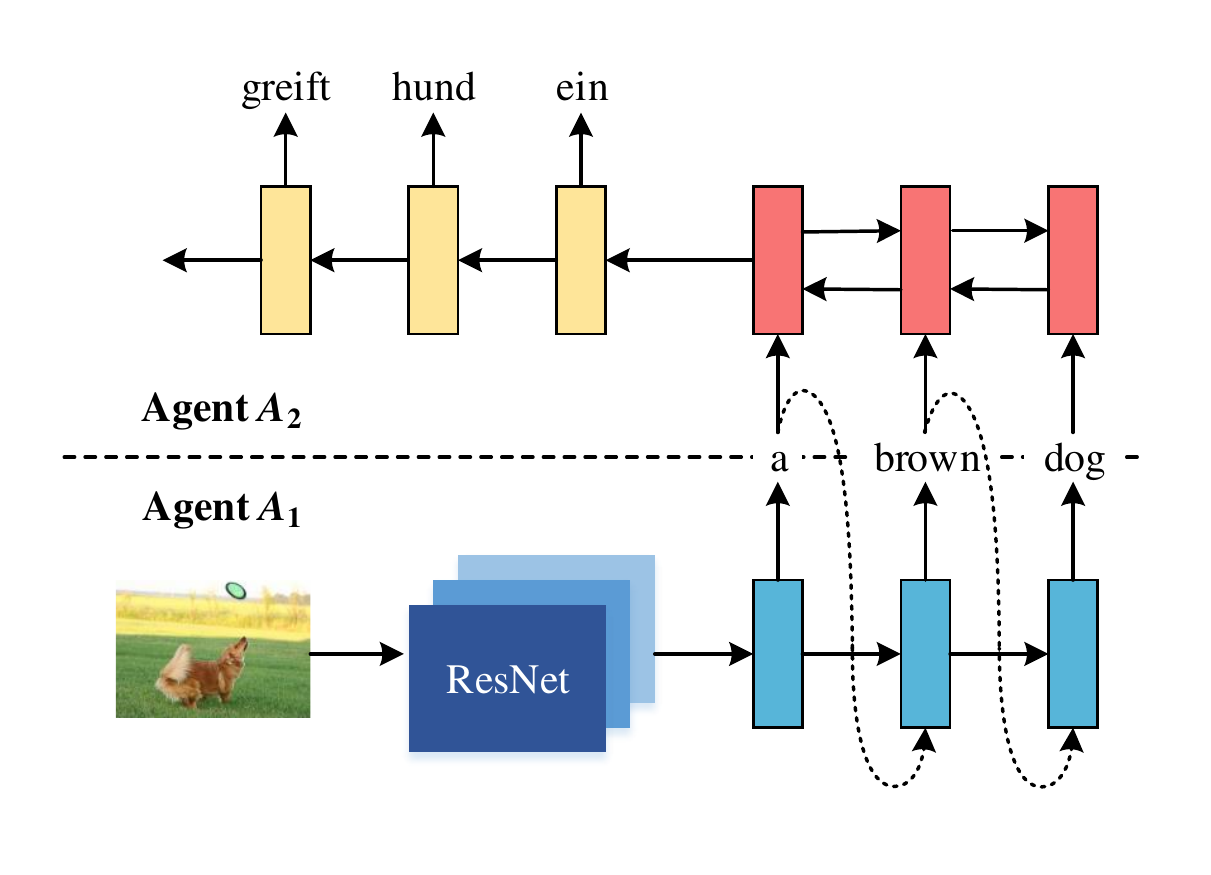}
	\caption{Zero-resource neural machine translation through a two-agent communication game within a multimodal environment. Agent $A_1$ is an image captioning model implemented with CNN-RNN architecture \cite{Xu2015ShowAA}; Agent $A_2$ is a neural machine translation model implemented with RNNSearch \cite{Bahdanau2014NeuralMT}. Taking an image as input, agent $A_1$ sends a message in the source language to agent $A_2$, which is translated by agent $A_2$ to a target-language sentence to win a reward.}\label{fig:sys}
\end{figure}
\subsection{Problem Formulation}
Given monolingual documents in the source language $D_{z,x} = \{ \langle \mathbf{z}^{(m)},\mathbf{x}^{(m)} \rangle \}_{m=1}^{M}$ and in the target language $D_{z,y} = \{ \langle \mathbf{z}^{(n)},\mathbf{y}^{(n)} \rangle \}_{n=1}^{N}$ , our aim is to learn a model which translates source sentence $\mathbf{x}$ to target sentence $\mathbf{y}$. Importantly, $D_{z,x}$ and $D_{z,y}$ do not overlap; they do not share the same images at all. Our model consists of a \emph{captioner} $P(\mathbf{x}|\mathbf{z}; \bm{\theta}_{z \rightarrow x})$ which translates image to source sentence and a \emph{translator} $P(\mathbf{y}|\mathbf{x}; \bm{\theta}_{x \rightarrow y})$ which translates source sentence to target sentence, where $\bm{\theta}_{z \rightarrow x}$ and $\bm{\theta}_{x \rightarrow y}$ are model parameters. To make the task a real zero-resource scenario, we assume that no parallel corpora are available even in the validation set. That is to say, we only have monolingual documents $D_{z,x}^{val}=\{ \langle \mathbf{z}^m,\mathbf{x}^m \rangle\}_{m=1}^{M^{val}}$ and $D_{z,y}^{val}=\{ \langle \mathbf{z}^n,\mathbf{y}^n  \rangle\}_{n=1}^{N^{val}}$ for validation.

In the following we describe two models: (i) the PRE. (pre-training) model that only trains the translator in the two-agent game as the captioner can be pre-trained with $D_{z,x}$; (ii) the JOINT model that jointly optimize the captioner and translator through reinforcement learning in the communication game. 

\subsection{The Game}
As illustrated in Fig. \ref{fig:sys}, we propose a simple communication game with two agents, the captioner $A_1$ and the translator $A_2$. Sampling a monolingual document $\langle \mathbf{z},\mathbf{y} \rangle$ from $D_{z,y}$, the game is defined as follows:
\begin{itemize}
 \item[1] $A_1$ is shown the image and is told to describe the image with a source-language sentence $\mathbf{x}_{mid} $.
 \item[2] $A_2$ is shown the middle sentence $\mathbf{x}_{mid}$ generated by $A_1$ without the image information. It is told to translate $\mathbf{x}_{mid}$ to a target-language sentence.
 \item[3] The environment evaluates the consistency of the translated target sentence and the gold-standard target-language sentence $\mathbf{y}$ and then both agents receive a reward.
\end{itemize}

The captioner $A_1$ and the translator $A_2$ must work together to achieve a good reward. $A_1$ should learn how to provide accurate image description in the source language and $A_2$ should be good at translating a source sentence to a target sentence. This game can be played for an arbitrary number of rounds, and the captioner $A_1$ and the translator $A_2$ will get trained through this reinforcement procedure (e.g., by means of the policy gradient methods). In this way, we develop a general learning framework for training zero-resource machine translation model (the translator $A_2$) with monolingual multimodal documents only through a multi-agent communication game. As we do not assume the specific architectures of the captioner and the translator, our proposed learning framework is transparent to architectures and can be applied to any end-to-end image captioning and NMT systems.

\subsection{Implementation}\label{sec:alg}
For a game beginning with a monolingual document $\langle \mathbf{z},\mathbf{y} \rangle \in D_{z,y}$, we use $\mathbf{x}_{mid}$ to denote the exchanged source sentence between agents. The goal of training is to find the parameters of the agents that maximize the expected reward:
\begin{eqnarray}
\label{eqn:goal}
\mathcal{E}(\bm{\theta}_{z \rightarrow x},\bm{\theta}_{x \rightarrow y})=\mathbb{E}_{P(\mathbf{x}_{mid}|\mathbf{z}; \bm{\theta}_{z \rightarrow x})}[r(\mathbf{y},\mathbf{x}_{mid},\bm{\theta}_{x \rightarrow y})]. 
\end{eqnarray}
We follow He et al. \shortcite{He2016DualLF} and define the reward as the log probability of agent $A_2$ generates $\mathbf{y}$ from $\mathbf{x}_{mid}$:
\begin{equation}
r(\mathbf{y},\mathbf{x}_{mid},\bm{\theta}_{x \rightarrow y})=\log P(\mathbf{y}|\mathbf{x}_{mid}; \bm{\theta}_{x \rightarrow y}).
\end{equation}
As a result, the expected reward in the multi-agent communication game can be re-written as:
\begin{equation}
\mathcal{E}(\bm{\theta}_{z \rightarrow x},\bm{\theta}_{x \rightarrow y})=\mathbb{E}_{P(\mathbf{x}_{mid}|\mathbf{z}; \bm{\theta}_{z \rightarrow x})}[\log P(\mathbf{y}|\mathbf{x}_{mid}; \bm{\theta}_{x \rightarrow y})]. 
\end{equation}

In training, we optimize the parameters of the captioner and translator through policy gradient methods for expected reward maximization:
\begin{eqnarray}
\hat{\bm{\theta}}_{z \rightarrow x},\hat{\bm{\theta}}_{x \rightarrow y} \qquad \qquad \qquad \qquad \qquad \quad \nonumber \\ = \argmax_{\bm{\theta}_{z \rightarrow x},\bm{\theta}_{x \rightarrow y}} \Big\{  \sum_{\langle \mathbf{z},\mathbf{y} \rangle \in D_{z,y} } \mathcal{E}(\bm{\theta}_{z \rightarrow x},\bm{\theta}_{x \rightarrow y}) \Big\}.
\end{eqnarray}
We compute the gradient of $\mathcal{E}(\bm{\theta}_{z \rightarrow x},\bm{\theta}_{x \rightarrow y})$ with respect to parameters $\bm{\theta}_{z \rightarrow x}$ and $\bm{\theta}_{x \rightarrow y}$. According to the policy gradient theorem \cite{Sutton1999PolicyGM}, it is easy to verify that:
\begin{eqnarray}
\label{eqn:grad1}
\nabla_{\bm{\theta}_{z \rightarrow x}} \mathcal{E}(\bm{\theta}_{z \rightarrow x},\bm{\theta}_{x \rightarrow y}) \qquad \qquad \nonumber \\ = \mathbb{E}[r \nabla_{\bm{\theta}_{z \rightarrow x}}\log P(\mathbf{x}_{mid}|\mathbf{z};\bm{\theta}_{z \rightarrow x})],
\end{eqnarray}
\begin{eqnarray}
\label{eqn:grad2}
\nabla_{\bm{\theta}_{x \rightarrow y}} \mathcal{E}(\bm{\theta}_{z \rightarrow x},\bm{\theta}_{x \rightarrow y}) \qquad \qquad \nonumber \\ =\mathbb{E}[\nabla_{\bm{\theta}_{x \rightarrow y}}\log P(\mathbf{y}|\mathbf{x}_{mid}; \bm{\theta}_{x \rightarrow y})],
\end{eqnarray}
in which the expectation is taken over $\mathbf{x}_{mid}$ and $r=\log[P(\mathbf{y}|\mathbf{x}_{mid}; \bm{\theta}_{x \rightarrow y})]$.

Unfortunately, Eqn. \ref{eqn:grad1} and \ref{eqn:grad2} are intractable to calculate due to the exponential
search space of $\cal X(\mathbf{z})$. Following \cite{He2016DualLF}, we adopt beam search for gradient estimation. Compared with random sampling, beam search can help to avoid the very large variance and sometimes unreasonable results brought by image captioning \cite{Ranzato2015SequenceLT}. Specifically, we run beam search with the captioner $A_1$ to generate top-$K$ high-probability middle outputs in the source language, and use the averaged value on the middle outputs to approximate the true gradient. Algorithm \ref{alg} shows the detailed learning algorithm.

\begin{algorithm}
	\caption{The learning algorithm in the two-agent communication game}\label{alg}
	\begin{algorithmic}[1]
		\State \textbf{Input}: Monolingual multimodal documents $D_{z,y}=\{\langle \mathbf{z}^n,\mathbf{y}^n \rangle \}_{n=1}^{N}$, initial image-to-source caption model $\theta_{z \rightarrow x}$, initial source-to-target translation model $\theta_{x \rightarrow y}$, beam search size $K$, learning rates $\gamma_{1,t},\gamma_{2,t}$.
		\Repeat
		\State $t=t+1$.
		\State Sample a document $\langle \mathbf{z}^n,\mathbf{y}^n \rangle$ from $D_{z,y}$.
		\State Set $\mathbf{z}=\mathbf{z}^n$, $\mathbf{y}=\mathbf{y}^n$. 
		\State Generate $K$ sentences $\mathbf{x}_{mid,1},\cdots ,\mathbf{x}_{mid,K}$ using beam search according to the image captioning model $P(\mathbf{x}|\mathbf{z}; \bm{\theta}_{z \rightarrow x})$.
		\For{$k=1,\cdots,K$}
		\State Set the reward for the $k$th sampled sentence as $r_k=\log P(\mathbf{y}|\mathbf{x}_{mid,k};\theta_{x \rightarrow y})$.
		\EndFor
		\State Compute the stochastic gradient of $\theta_{z \rightarrow x}$:
		\begin{eqnarray}
		\nabla_{\bm{\theta}_{z \rightarrow x}} \hat{\mathbb{E}}[r] = \frac{1}{K}\sum_{k=1}^K[r_k \nabla_{\bm{\theta}_{z \rightarrow x}}\log P(\mathbf{x}_{mid,k}|\mathbf{z}; \bm{\theta}_{z \rightarrow x})].\nonumber
		\end{eqnarray}
		\State Compute the stochastic gradient of $\theta_{x \rightarrow y}$:
		\begin{eqnarray}
		\nabla_{\bm{\theta}_{x \rightarrow y}} \hat{\mathbb{E}}[r] = \frac{1}{K}\sum_{k=1}^K[ \nabla_{\bm{\theta}_{x \rightarrow y}}\log P(\mathbf{y}|\mathbf{x}_{mid,k}; \bm{\theta}_{x \rightarrow y})].\nonumber
		\end{eqnarray}
		\State Model updates:
		\begin{eqnarray}
		\theta_{z \rightarrow x} \leftarrow\theta_{z \rightarrow x} + \gamma_{1,t}\nabla_{\theta_{z \rightarrow x}} \hat{\mathbb{E}}[r], \nonumber \\
		\theta_{x \rightarrow y} \leftarrow\theta_{x \rightarrow y} + \gamma_{2,t}\nabla_{\theta_{x \rightarrow y}} \hat{\mathbb{E}}[r].\nonumber 
		\end{eqnarray}
		\Until{convergence}
	\end{algorithmic}
\end{algorithm}
 
\subsection{Training}
Since the captioner $A_1$ has a very large action space to generate source description $\mathbf{x}_{mid}$, it is extremely difficult to learn with an initial random policy. Specifically, the search space for $A_1$ is of size $\mathcal{O}(\mathcal{V}^T )$, where $\mathcal{V}$ is the number of words in the source vocabulary (more than 1000 in our experiments) and $T$ is the length of the sentence (around 10 to 20 in our experiments). Thus, we pre-train the captioner $A_1$ with maximum likelihood estimation leveraging monolingual dataset $D_{z,x}$:
\begin{equation}
\hat{\bm{\theta}}_{z \rightarrow x}^{\mathrm{pre}}
=\argmax_{\bm{\theta}_{z \rightarrow x}}\Bigg\{\sum_{\langle \mathbf{z},\mathbf{x} \rangle \in D_{z,x} } \log P(\mathbf{x}|\mathbf{z}; \bm{\theta}_{z \rightarrow x})\Bigg\}. 
\end{equation}
We initialize the captioner $A_1$ with the pre-trained image caption model and randomly initialize the translator $A_2$. To avoid the randomly initialized agent $A_2$ doing harm to the pre-trained captioner $A_1$, we fix $A_1$ for the initial few epochs and only optimize agent $A_2$. Then we adopt two different training approaches: 
\begin{itemize}
	\item[1.] The PRE. (pre-training) model: We keep the captioner fixed and only train the translator in the two-agent game. Thus, the training objective is:
	\begin{eqnarray}
	\mathcal{J}_{\mathrm{PRE.}}(\bm{\theta}_{x \rightarrow y}) \quad \quad \quad  \quad \quad \quad \quad \quad \quad \quad \quad \quad \nonumber \\
	=\sum_{\langle \mathbf{z},\mathbf{y} \rangle \in D_{z,y} } \mathbb{E}_{P(\mathbf{x}_{mid}|\mathbf{z}; \hat{\bm{\theta}}_{z \rightarrow x}^{\mathrm{pre}})}[\log P(\mathbf{y}|\mathbf{x}_{mid}; \bm{\theta}_{x \rightarrow y})]. 
	\end{eqnarray}
	
	\item[2.] The JOINT model: We jointly optimize the captioner and the translator through reinforcement learning in the communication game. To encourage the captioner $A_1$ to correctly describe the image in the source language, we use half documents from monolingual dataset $D_{z,x}=\{ \langle \mathbf{z}^{(m)},\mathbf{x}^{(m)} \rangle \}_{m=1}^{M}$ in each mini batch to constrain the parameters of $A_1$. We train to maximize the weighted sum of the reward based on documents from $D_{z,y}$ and the log-likelihood of image-to-source captioning model $\theta_{z \rightarrow x}$ on documents from $D_{z,x}$. The objective becomes:
	\begin{multline}
	\quad \quad \quad \mathcal{J}_{\mathrm{JOINT}}(\bm{\theta}_{z \rightarrow x},\bm{\theta}_{x \rightarrow y}) \\
	=\sum_{\langle \mathbf{z},\mathbf{y} \rangle \in D_{z,y} } \mathbb{E}_{P(\mathbf{x}_{mid}|\mathbf{z}; \bm{\theta}_{z \rightarrow x})}[\log P(\mathbf{y}|\mathbf{x}_{mid}; \bm{\theta}_{x \rightarrow y})]\\
	+\lambda\sum_{\langle \mathbf{z},\mathbf{x} \rangle \in D_{z,x} } \log P(\mathbf{x}|\mathbf{z}; \bm{\theta}_{z \rightarrow x}). \quad \quad \quad
	\end{multline}
\end{itemize}

\subsection{Validation}
\begin{figure}[!t]
	\centering\includegraphics[width=2.8in]{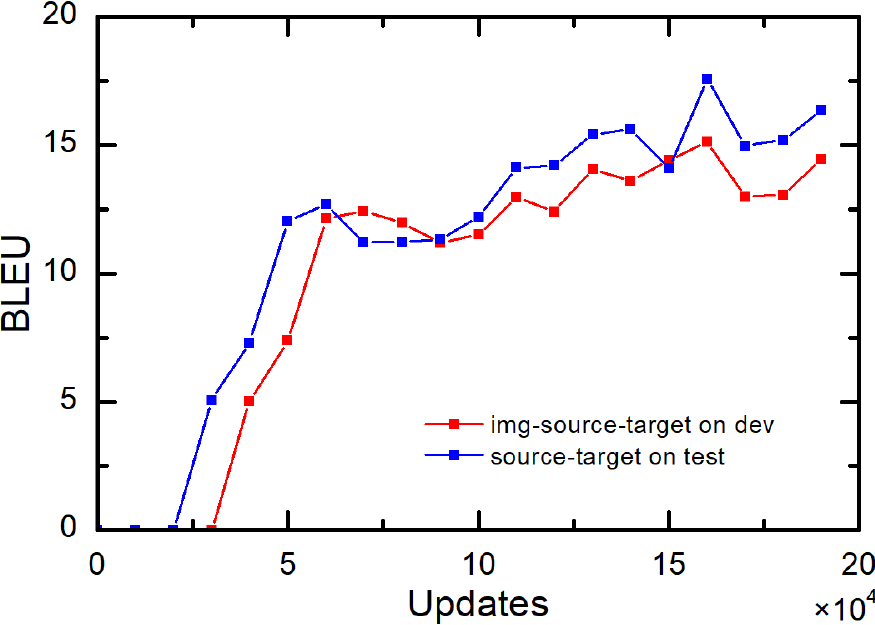}
	\caption{Validation of the  PRE. model on monolingual documents $D_{z,y}^{val}$ for the IAPR-TC12 German-to-English translation task. BLEU score of the translator $A_2$ on the test set correlates very well with BLEU score on $D_{z,y}^{val}$ using our proposed validation criterion. }\label{fig:valid}
\end{figure}
Since we do not have access to parallel sentences even at validation time, we need to have a criterion for model and  hyper-parameters selection for the PRE. and JOINT methods. We validate the model on $D_{z,y}^{val}$. 

Given model parameters $\bm{\theta}_{z \rightarrow x}$ and $\bm{\theta}_{x \rightarrow y}$ at some iteration and a monolingual document $\langle \mathbf{z},\mathbf{y} \rangle$ from $D_{z,y}^{val}$, we output the image's description in the target language with the following decision rule using beam search:
\begin{eqnarray}
\hat{\mathbf{x}} = \argmax_{\mathbf{x}}\Big\{ P(\mathbf{x}|\mathbf{z}; \bm{\theta}_{z \rightarrow x}) \Big\} \\
\hat{\mathbf{y}} = \argmax_{\mathbf{y}}\Big\{ P(\mathbf{y}|\hat{\mathbf{x}}; \theta_{x \rightarrow y}) \Big\}
\end{eqnarray}
With the generated $\hat{\mathbf{y}}$ as the hypothesis and the gold-standard description $\mathbf{y}$ as the reference, we use the BLEU score as the validation criterion, namely, we choose the translation model with the highest BLEU score. Figure \ref{fig:valid} shows a typical example of the correlation between this measure and the translation performance of the translator $A_2$ on test set.

\section{Experiments}
\subsection{Data Set}
\begin{table}[t!]
	\caption{\label{table:stat1}Dataset statistics.}
	\begin{center}
		\begin{tabular}{|l|l|c|c|}
			\hline
			\multicolumn{2}{|c|}{} & IAPR-TC12 & Multi30K \\
			\hline
			\multicolumn{2}{|c|}{ Num. of images} & 20,000 & 31,014\\ \hline
			\multirow{2}{*}{Vocabulary} & En & 1207 &  5877 \\
			& De & 1511 &  7225 \\ \hline
			Avg. length    & En & 18.8 & 14.1 \\ 
			of descriptions & De & 15.9 & 11.7 \\\hline 
		\end{tabular} 
	\end{center}
\end{table}
\begin{table*}
	\caption{\label{table:stat2}Splits for experiments. 
		Each image is annotated with one English or German sentence for the IAPR-TC12 dataset, while each image is described by 5 English or German sentences for the Multi30K dataset. }
	\begin{center}
		\begin{tabular}{|l|c|c|c|c|c|c|c|}
			\hline
			\multirow{2}{*}{Split} & \multirow{2}{*}{Pair} & \multicolumn{3}{|c|}{IAPR-TC12} & \multicolumn{3}{|c|}{Multi30K} \\
			\cline{3-8}
			& & img & En & De & img & En & De \\
			\hline
			\multirow{2}{*}{Train} & img-En & 9,000 & 9,000 & - & 14,500 & 72,500 & - \\
			& img-De & 9,000 & - & 9,000 & 14,500 & - & 72,500 \\ \hline
			\multirow{2}{*}{Validation} & img-En & 500 & 500 & - & 507 & 2,535 & - \\
			& img-De & 500 & - & 500 & 507 & - & 2,535 \\ \hline
			Test & En-De & - & 1,000 & 1,000 & - & 5,000 & 5,000 \\ \hline
		\end{tabular} 
	\end{center}
\end{table*}
We evaluate our model on two publicly available multilingual image-description datasets as in \cite{Nakayama2017ZeroresourceMT}. The IAPR-TC12 dataset \cite{Grubinger2006TheIT}, which consists of English image descriptions and the corresponding German translations, has a total of 20K images. Each image contains multiple descriptions and each description corresponds to a different aspect of the image. Since the first sentence is likely to describe the most salient objects \cite{Grubinger2006TheIT}, we use only the first description of each image. We randomly split the dataset into training, validation and test sets with 18K, 1K and 1K images respectively. The recently published Multi30K dataset \cite{elliott-EtAl:2016:VL16}, which is a multilingual extension of Flickr30k corpus \cite{young2014image}, has 29K, 1K and 1K images in the training, validation and test splits respectively with English and German image descriptions \cite{elliott-EtAl:2016:VL16}. There are two types of multilingual annotations in the dataset: (i) a corpus of one English description per image and its German translation; and (ii) a corpus of 5 independently collected English and German descriptions per image. Since the corpus of independently collected English and German descriptions better fit the noisy multimodal content on the web, we adopt this corpus in our experiments. Note that although these descriptions describe the same image, they are not translations of each other. 

For preprocessing, we use the scripts in the Moses SMT
Toolkit \cite{Koehn2007MosesOS} to normalise and
tokenize English and German descriptions. For the IAPR-TC12 dataset, we construct the vocabulary with words appearing more than 5 times in the training splits and replace those appearing less than 5 times with UNK symbol. 
For the Multi30K dataset, we adopt a joint byte pair encoding (BPE) \cite{Sennrich2016NeuralMT} with 10K merge operations on English and German descriptions to reduce vocabulary size. 
To comply with the zero-resource setting, we randomly split the images in the training and validation datasets into two parts with equal size. One part constructs the image-English split and the other part the image-German split. Unnecessary modalities for each split (e.g., German descriptions for image-English split) are ignored. Note that the two splits have no overlapping images, and we have no direct English-German parallel corpus. Table \ref{table:stat1} and \ref{table:stat2} summarizes data statistics.
\subsection{Experimental Setup}
To extract image features, we follow the suggestion of \cite{Caglayan2016DoesMH} and adopt ResNet-50 network \cite{He2016DeepRL} pre-trained on ImageNet without finetuning. We use the (14,14,1024) feature map of the res4fx (end of Block-4) layer after ReLU. For some baseline methods that do not support attention mechanism, we extract 2048-dimension feature after the pool5 layer. We follow the architecture in \cite{Xu2015ShowAA} for image captioning and standard RNNSearch architecture \cite{Bahdanau2014NeuralMT} for translation. We leverage {\em dl4mt} \footnote{\textit{dl4mt-tutorial}: https://github.com/nyu-dl} and {\em arctic-captions} \footnote{\textit{arctic-captions}: https://github.com/kelvinxu/arctic-captions} for all our experiments. The beam search size is 2 in the middle image-to-source caption generation. During validation and testing, we set the beam search size to be 5 for both the captioner and the translator. All models are quantitatively evaluated with BLEU \cite{papineni2002bleu}. For the Multi30K dataset, each image is paired with 5 English descriptions and 5 German descriptions in the test set. We follow the setting in \cite{Caglayan2016DoesMH} and let the NMT generate a target description for each of the 5 source sentences and pick the one with the highest probability as the translation result. The evaluation is performed against the corresponding five target descriptions in the testing phase. 

We compare our approach with four baseline methods:
\begin{itemize}
	\item[1] Random: For a sentence in source language, we randomly select a document in $D_{z,y}$ whose caption would be output as the translation result.
	\item[2] TFIDF: For a source sentence, we first search the nearest document in $D_{z,x}$ in terms of cosine similarity of TFIDF text features. Then, for the coupled image of that document, we retrieve the most similar document in $D_{z,y}$ in terms of cosine similarity of pool5 feature vectors. We output the coupled target sentence of the retrieved document as our translation.
	\item[3] 3-way model \cite{Nakayama2017ZeroresourceMT}: We leverage the best 3-way model proposed in \cite{Nakayama2017ZeroresourceMT} as our baseline, which adopts end to end training strategy and trains the decoder with image and description.
	\item[4] TS model \cite{Chen2017ATF}: This method is originally designed for leveraging a third language as a pivot to enable zero-resource NMT. We follow the teacher-student framework and replace the pivot language with image. The teacher model is an image-to-target captioning model and the student model is the zero-resource source-to-target translation model.
\end{itemize}

\subsection{Comparison with Baselines}
\begin{table}[!t]	
	\caption{\label{table:rs1}Comparison with previous work on German-to-English and English-to-German translation with zero resource over the IAPR-TC12 and Multi30K datasets.}
	\begin{center}
		\begin{tabular}{|l|c c|c c|}
			\hline
			& \multicolumn{2}{|c|}{IAPR-TC12} & \multicolumn{2}{|c|}{Multi30K} \\ \cline{2-5}
			&  De-En & En-De  & De-En & En-De \\
			\hline 
			Random  & 2.5 & 1.2 & 2.0 & 0.8 \\
			TFIDF & 10.2 & 7.5 & 2.4 & 1.0 \\
			3-way model & 13.9 & 8.6 & 15.9 & 10.1 \\ 
			TS model       & 14.1 & 9.2 & 16.4 & 10.3 \\ \hline
			PRE.     & 17.6 & 11.9 & 16.0 & 12.1 \\
			JOINT      & \textbf{18.6} & \textbf{14.2} & \textbf{19.6} & \textbf{16.6} \\\hline
		\end{tabular} 
	\end{center}
\end{table}
\begin{figure*}[!t]
	\centering\includegraphics[width=6.7in]{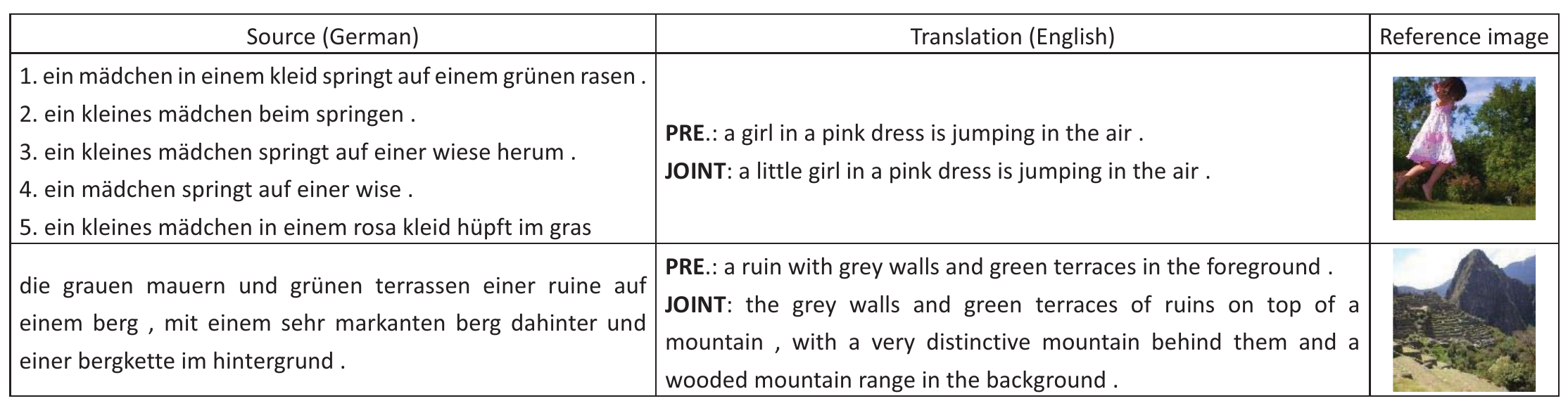}
	\caption{Examples of target translations from the test set using zero-resource NMT trained by our proposed PRE. and JOINT models. The first example is from the Multi30K dataset, while the second is from the IAPR-TC12 dataset.}\label{fig:example}
\end{figure*}
Table \ref{table:rs1} shows the translation performance of our proposed zero-resource NMT on the IAPR-TC12 and Multi30K datasets in comparison with other baselines. Note that the TS model, PRE. model and JOINT model all utilize attention mechanism, while the 3-way model does not support attention. It is clear from the table that in all the cases, our proposed JOINT model outperforms all the other baselines. Even with fixed captioner, the PRE. model can outperform the baselines in most cases.

Specifically, the JOINT model outperforms the 3-way model by +4.7 BLEU score on German-to-English translation and +5.6 BLEU score on English-to-German translation over the IAPR-TC12 dataset; +3.7 BLEU score on German-to-English translation and +6.5 BLEU score on English-to-German translation for the Multi30K dataset. The performance gap can be explained since the 3-way model attempts to encode a whole input sentence or image into a single fixed-length vector, rendering it difficult for the neural network to cope with long sentences or images with a lot of clutter. In contrast, the JOINT model adopts the attention mechanism for the captioner and the translator. Rather than compressing an entire image or source sentence into a static representation, attention allows for a model to automatically attend to parts of a source sentence or image that are relevant to predicting a target-side word. This explanation is in line with the observation that all three models with attention mechanism outperform the 3-way model.

Although the TS model also adopts the attention mechanism, its performance is dominated by the performance of the teacher model, namely the image captioning model. During the learning process of the student model, the teacher model is kept fixed all the time and the student tries to mimic the decoding behavior of the teacher in the teacher-student framework. The description of the same image can vary a lot in different languages as indicated in \cite{Miltenburg2017CrosslinguisticDA}. Thus, the teacher model's captioning result in target language is not necessarily the translation of the image's coupled source-language description. On the contrary, the JOINT model jointly optimize the captioner and the translator to win a two-agent communication game. The two agents are complementary to each other and are trained jointly to maximize expected reward. This mechanism helps to solve the problem of cross-linguistic differences in image description \cite{Miltenburg2017CrosslinguisticDA}, resulting in better zero-resource NMT model.

Figure \ref{fig:example} shows translation examples of the zero-resource NMT trained by the PRE. and JOINT methods. We also list the corresponding image for reference. The first example is from the Multi30K dataset, while the second is from the IAPR-TC12 dataset. Apparently, the JOINT model has successfully learned how to translate even with zero-resource.

\subsection{Effect of Joint Training}
\begin{table}[!t]	
	\caption{\label{table:joint}Comparison the captioner's performance of our proposed PRE. and JOINT model.}
	\begin{center}
		\begin{tabular}{|l|c c |c c |}
			\hline
			\multirow{2}{*}{Model}& \multicolumn{2}{|c|}{IAPR-TC12} & \multicolumn{2}{|c|}{Multi30K} \\ \cline{2-5}
			&  img-de & img-en & img-de & img-en \\
			\hline
			PRE. & 10.7 & 18.0 & 10.9 & 22.7 \\
			JOINT & 10.5 & 17.3 & 12.3 & 21.3  \\
			\hline
		\end{tabular} 
	\end{center}
\end{table}
\begin{figure*}[!t]
	\centering\includegraphics[width=6.3in]{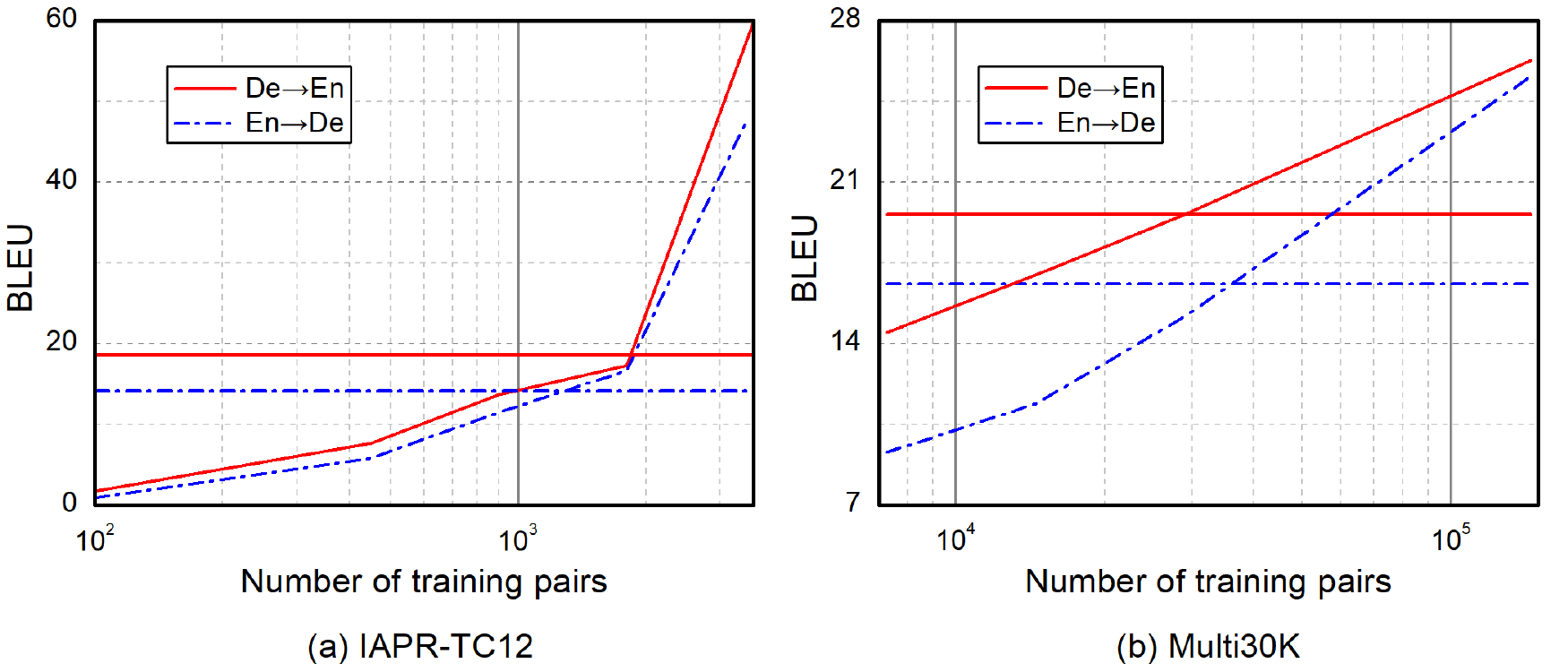}
	\caption{Comparison with ORACLE that uses direct source-target parallel or comparable corpora. The JOINT model corresponds to horizontal lines.}\label{fig:oracle}
\end{figure*}
We also compare the performance of the captioner with (JOINT) and without (PRE.) the joint training. Table \ref{table:joint} shows the result. Cheng et al. \shortcite{Cheng2017JointTF} demonstrate that in pivot-based NMT, where the pivot is a third language, the source-pivot and pivot-target translation models can be improved simultaneously through joint training. In our setting, image-to-source captioning does not necessarily improve with joint training. It gets slightly worse on three out of four translation tasks. Using a third language as a pivot, the source-to-pivot and pivot-to-target models are symmetrical as they are both NMT models. In contrast, in our setting the captioner and the translator are not symmetrical anymore since NMT is much easier to improve than image caption. We suspect that the translator's performance dominates the multi-agent game.

\subsection{Comparison with Oracle} 
Figure \ref{fig:oracle} compares the JOINT model with ORACLE that uses direct source-target parallel or comparable corpora on the IAPR-TC12 and Multi30K dataset. 

For the Multi30K dataset, each image is annotated with 5 German sentences and 5 English sentences. The training dataset for ORACLE can be constructed by the cross product of 5 source and 5 target descriptions which results in a total of 25 description pairs for each image or by only taking the 5 pairwise descriptions. We follow \cite{Caglayan2016DoesMH} and use the pairwise way.

For the IAPR-TC12 dataset, the JOINT model is roughly comparable with ORACLE when the number of parallel sentences are limited to about 15\% that of our monolingual ones. For the Multi30K dataset, the JOINT model obtains comparable results with ORACLE trained by number of comparable sentence pairs about 40\% that of our monolingual corpus. It obtains 75\% of the BLEU score on German-to-English translation and 65\% of the BLEU score on English-to-German translation for NMT trained with full corpora. Although there is  still a significant gap in performance as compared to ORACLE trained with parallel corpus, this is encouraging since our approach only uses monolingual multimodal documents.

\section{Related Work}
Training NMT models without source-target parallel corpora by leveraging a third language or image modality has attracted intensive attention in recent years. Utilizing a third language as a pivot has already achieved promising translation quality for zero-resource NMT. Firat et al. \shortcite{Firat2016ZeroResourceTW} pre-train multi-way multilingual model and then fine-tune the attention mechanism with pseudo parallel data generated by the model to improve zero-resource translation. Johnson et al. \cite{Johnson2016GooglesMN} adopt a universal encoder-decoder network in multilingual scenarios to naturally enable zero-resource translation. In addition to the above multilingual methods, several authors propose to train the zero-resource source-to-target translation model directly. Chen et al. \shortcite{Chen2017ATF} propose a teacher-student framework under the assumption that parallel sentences have close probabilities of generating a sentence in a third language. Zheng et al. \shortcite{Zheng2017MaximumEL} maximize the expected likelihood to train the intended source-to-target model. However, all these methods assume that source-pivot and pivot-target parallel corpora are available. Another line is to bridge zero-resource language pairs via images. Nakayama and Nishida \shortcite{Nakayama2017ZeroresourceMT} train multimodal encoders to learn modality-agnostic multilingual representation of fix length using image as a pivot. On top of the fix-length representation, they build a decoder to output a translation in the target language. Although the performance is limited by the fix-length representation, their work shows that zero-resource neural machine translation with an image pivot is possible.

Multimodal neural machine translation, which introduce image modality into NMT as an additional information source to reinforce the translation performance, has received much attention in the community recently. A lot of work have shown that image modality can benefit neural machine translation, hopefully by relaxing ambiguity in alignment that cannot be solved by texts only \cite{Hitschler2016MultimodalPF,Calixto2017DoublyAttentiveDF}. Note that their setting is much easier than ours because in their setting multilingual descriptions for the same images are available in the training dataset and an image is part of the query in both training and testing phases.

\section{Conclusion}
In this work, we propose a multi-agent communication game to tackle the challenging task of training a zero-resource NMT system from just monolingual multimodal data. In contrast with previous studies that learn a modality-agnostic multilingual representation, we successfully deploy the attention mechanism to the target zero-resource NMT model by encouraging the agents to cooperate with each other to win a image-to-target translation game. Experiments on German-to-English and English-to-German translation over the IAPR-TC12 and Multi30K datasets show that our proposed multi-agent learning mechanism can significantly outperform the state-of-the-art methods.

In the future, we plan to explore whether machine translation can perform satisfactorily with automatically crawled noisy multimodal data from the web. Since our current method is intrinsically limited to the domain where texts can be grounded to visual content, it is also interesting to explore how to further extend the learned translation model to handle generic documents. 

\section*{Acknowledgments}
This work is partially supported by the National Key R\&D Program of China (No. 2017YFB0202204), National Natural Science Foundation of China (No. 61522204, No. 61331013) and the HKU Artificial Intelligence to Advance Well-being and Society (AI-WiSe) Lab.

\bibliographystyle{aaai} 
\bibliography{aaai18}

\end{document}